\documentclass{article} 
\usepackage{iclr2025_conference,times}

\usepackage{arxiv}

\usepackage[utf8]{inputenc} 
\usepackage[T1]{fontenc}    
\usepackage{hyperref}
\usepackage{url}            
\usepackage{booktabs}       
\usepackage{amsfonts}       
\usepackage{nicefrac}       
\usepackage{microtype}      
\usepackage{xcolor}         
\usepackage{etoc}           
\usepackage{pifont}
\usepackage{adjustbox}
\usepackage{wrapfig}
\usepackage{multirow,mathtools}
\usepackage{rotating}
\usepackage{lscape}
\usepackage{longtable}
\usepackage[font={small}]{caption}
\usepackage{amssymb}
\usepackage{fontawesome}
\usepackage{graphicx}
\graphicspath{{figures/}}

\newcommand{\cmark}{\textcolor{green}{\ding{51}}}
\newcommand{\xmark}{\textcolor{red}{\ding{55}}}

\usepackage{colortbl}
\definecolor{Gray}{gray}{0.93}
\definecolor{Orange}{rgb}{1,0.5,0}
\definecolor{DGray}{gray}{0.83}
\definecolor{LightCyan}{rgb}{0.88,1,1}

\definecolor{darkgreen}{rgb}{0.0, 0.65, 0.0}
\definecolor{darkred}{rgb}{0.5, 0.0, 0.0}
\definecolor{darkblue}{rgb}{0.0, 0.0, 0.5}
\definecolor{darkyellow}{rgb}{0.65, 0.65, 0}

\definecolor{mr}{RGB}{251, 111, 111}

\usepackage[english]{babel}

\usepackage{amsthm}

\usepackage[capitalize,noabbrev]{cleveref}

\theoremstyle{plain}

\theoremstyle{definition}

\theoremstyle{remark}

\usepackage{tcolorbox}
\usepackage{enumitem}

\usepackage{bm}

\def\eqref#1{(\ref{#1})}

\def\1{\bm{1}}

\def\vr{{\bm{r}}}

\DeclareMathAlphabet{\mathsfit}{\encodingdefault}{\sfdefault}{m}{sl}
\SetMathAlphabet{\mathsfit}{bold}{\encodingdefault}{\sfdefault}{bx}{n}

\input{preamble}

\title{
\ours: Learning What Changes, Not What Stays, for World-Action Models
}

\author{Chongyu Fan$^{\dag}$ ~~Gaowen Liu$^{\ddag}$ ~~Ramana Rao Kompella$^{\ddag}$ ~~Jason Corso$^{\P}$ ~~Sijia Liu$^{\dag,\S}$\\
  $^\dag$Michigan State University ~~$^\ddag$Cisco ~~$^\P$University of Michigan ~~$^\S$IBM Research
}
\date{}

\begin{document}

\pagestyle{fancy}
\fancyhf{}
\cfoot{\thepage}

\maketitle

\vspace{-14mm}
\begin{center}
\href{https://chongyu-fan.netlify.app/posts/reswam/}{\faGlobe\ \texttt{Project Page}} \quad\textbar\quad
\href{https://github.com/OPTML-Group/ResWAM}{\faGithub\ \texttt{Code}} \quad\textbar\quad
\href{https://huggingface.co/collections/OPTML-Group/reswam-6af8906dc9b6f21539b30934}{\faDatabase\ \texttt{Models}}
\end{center}
\vspace{2mm}

\begin{abstract}
World-action models (WAMs) augment vision-language-action policies (VLAs) with future prediction so that the policy can anticipate how its actions will affect the physical world. Existing WAMs supervise complete future observations. However, observations typically change little over an action chunk, so the future loss is dominated by copying static background content rather than learning task-relevant robot and object motion, limiting policy learning. Moreover, existing WAMs rely on large video generation models or separate world models, incurring heavy training and inference compute that hinders closed-loop deployment. To improve policy performance, we propose \emph{visual residual learning}, supervising only the feature change from the current frame to the future frame in a frozen DINOv3 space. Compared with complete future features, this residual learning target yields a higher feature rank with richer and more diverse representations of scene changes, and spatially concentrates on the robot arm and manipulated objects rather than the background. To further improve computational efficiency, we present \ours, a residual world-action model with 2.3B parameters in total, including a lightweight 0.5B dual-branch DiT that jointly trains an action branch and a residual branch via conditional flow matching. Without robot pretraining (\textit{i.e.}, pretraining on external robot action trajectories beyond the benchmark demonstrations), \ours achieves top average success rates on LIBERO-Plus (81.1\%), RoboTwin~2.0 (93.8\%), and RoboDojo (9.40\% success, 13.41 score) at the lowest training (2.7--4.0\,TFLOPs) and inference compute (1.0--1.4\,TFLOPs), and leads real-robot Open, Put, and Stack (80.0\% average, $+10.0\%$ over $\pi_{0.5}$) at 92.7\,ms latency.
\end{abstract}
\begin{figure}[htb]
  \centering
  \includegraphics[width=0.97\textwidth]{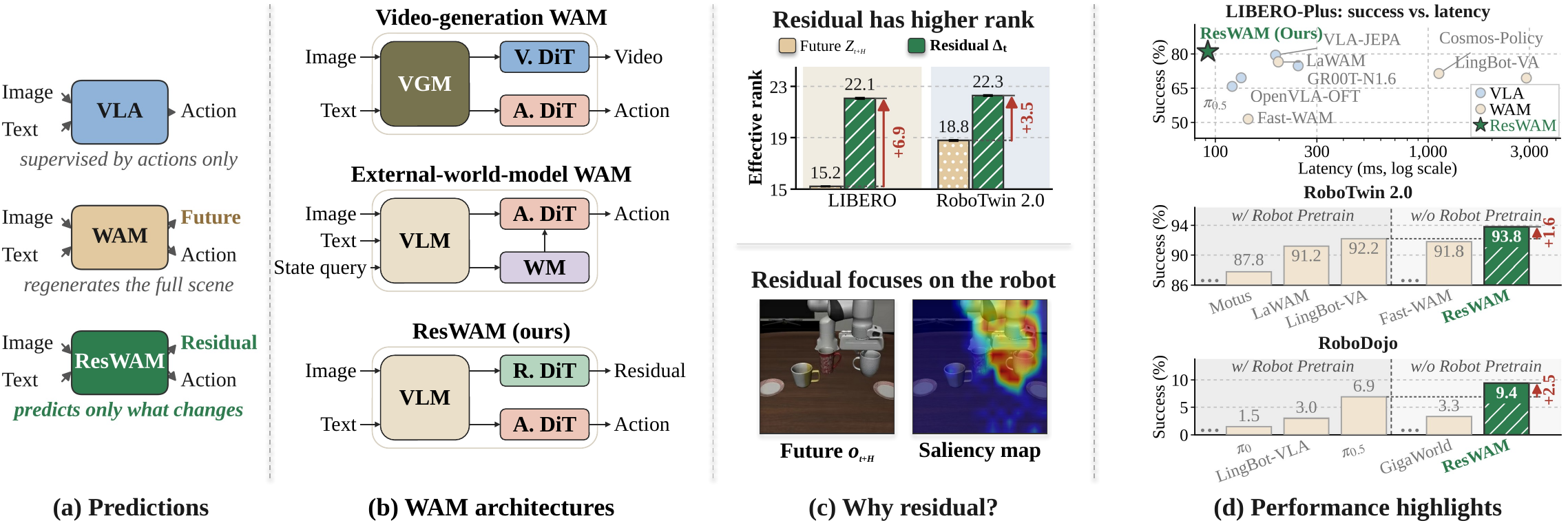}
  \caption{\footnotesize
  Schematic overview of \ours across (a) predictions, (b) architectures, (c) benefits of residuals, and (d) performance highlights.
  (a) Predictions of VLAs, WAMs, and \ours.
  (b) Architectures of WAMs: video-generation WAMs use a video generation model (VGM);
  external-world-model WAMs use a vision-language model (VLM) and separate
  world model (WM); \ours is built on a VLM with residual and action branches.
  V.\ DiT, A.\ DiT, and R.\ DiT denote video, action, and residual Diffusion
  Transformers, respectively.
  (c) Residual features have higher effective rank and focus on the robot and
  manipulated objects.
  (d) LIBERO-Plus success versus latency, and RoboTwin~2.0 and RoboDojo average success rates across methods. Here w/ and w/o Robot Pretrain indicate with and without robot pretraining on external robot action datasets prior to downstream training.}
  \label{fig:teaser}
\end{figure}

\section{Introduction}
\label{sec:intro}

As frontier AI models advance rapidly, physical intelligence is emerging as a key frontier for perceiving, reasoning about, and interacting with the physical world~\cite{seed2026seed2,fan2026cyclicreflex}.
Current approaches broadly follow two paradigms: vision-language-action models (VLAs) and world-action models (WAMs), shown in \textbf{\cref{fig:teaser}(a)}.
VLAs adapt pretrained vision-language models (VLMs) to predict actions from visual and language inputs~\cite{black2024pi0,black2025pi0,fan2026up,kim2025fine}, but train only on action labels without modeling how actions change the physical world.

WAMs add future visual prediction alongside actions so the policy accounts for the physical consequences of its commands~\cite{bi2025motus,li2026causal,wang2026openwam,yang2026lila,yuan2026fast}. Existing WAMs typically realize this by predicting complete future frames or features via a video generation model or a separate world model (\textbf{\cref{fig:teaser}(b)}). However, this standard paradigm suffers from \textit{two fundamental limitations} in \emph{supervision target} and \emph{model architecture}.

First, in terms of \emph{supervision target}, supervising complete future observations weakens policy learning because consecutive frames over an action chunk change very little. For instance, on LIBERO~\cite{liu2023libero}, current and future visual features share an average cosine similarity of 0.96; consequently, the future loss is largely satisfied by copying static background tokens rather than capturing how the robot and objects move in response to actions, diluting the auxiliary supervision signal for control. To mitigate this issue, recent studies have explored compact representations of temporal changes.
Yet, these approaches still rely on indirect modeling or additional architectural complexity, such as latent-action encoding and future-feature reconstruction~\cite{zhang2026learning}, action inference from latent transitions~\cite{qiu2026ad,zhang2026delta}, or delta tokenization coupled with video generation~\cite{yan2026deltawam}.
Second, in terms of \emph{model architecture}, existing WAMs incur high training and inference cost from large video generation models or separate world models (\cref{fig:teaser}(b)). Video-generation WAMs~\cite{bi2025motus,kim2026cosmos,li2026causal,yuan2026fast} fine-tune large video diffusion models with costly 3D video VAEs, while external-world-model WAMs~\cite{chen2026lawam,lin2026jepa,sun2026vla,yang2026lila} require separate pretrained world models or multi-stage teachers.

The above limitations (static-background dominance in supervision and heavy backbone cost in architecture) motivate our hypothesis:
\begin{tcolorbox}[before skip=1.4mm, after skip=1.4mm, boxsep=0.0cm, left=1.5mm, right=1.5mm, middle=0.0cm, top=0.05cm, bottom=0.05cm, boxrule=0.6pt]
\begin{center}
    \textit{\textbf{(H)} Supervising a lightweight DiT with visual residuals in WAMs can achieve improved performance while reducing computational costs.}
\end{center}
\end{tcolorbox}

To approach \textbf{(H)}, we introduce \textit{visual residual learning}, which shifts WAM supervision from predicting \emph{what the future world looks like} to predicting \emph{how the world changes}. Given visual tokens $\Lat_t$ and $\Lat_{t+H}$ extracted from the current frame and the frame $H$ steps ahead using a frozen DINOv3~\cite{simeoni2025dinov3} encoder, respectively, we construct the visual residual target as $\Res_t=\Lat_{t+H}-\Lat_t$, capturing temporal changes while suppressing static background content. We identify two key advantages of this supervision strategy, as illustrated in \textbf{\cref{fig:teaser}(c)}. \textit{First}, visual residuals exhibit higher feature rank than future representations, suggesting richer and more diverse information (\textbf{\cref{sec:rank}}). \textit{Second}, visualizing the residuals in image space reveals that they concentrate on the robot and manipulated objects rather than static backgrounds, highlighting action-relevant scene changes (\textbf{\cref{sec:vis}}). Empirically, supervising $\Res_t$ improves average success rate by $+2.9\%$ over directly predicting $\Lat_{t+H}$ at identical compute on RoboTwin~2.0 (\textbf{\cref{fig:ablation}}).

On the architectural side, we design {\ours} (\cref{fig:teaser}(b)) to improve computational efficiency. Instead of using a large video generation model or a separate world model, {\ours} has only 2.3B parameters in total, including a lightweight 0.5B dual-branch DiT that jointly denoises actions and frozen DINOv3 visual residuals via conditional flow matching, without pixel reconstruction or future decoding. Across all evaluated benchmarks, {\ours} achieves the lowest training compute (2.7--4.0\,TFLOPs), inference compute (1.0--1.4\,TFLOPs), and inference latency (92.2--99.7\,ms). \textbf{\cref{fig:teaser}(d)} highlights the favorable success-latency trade-off of {\ours} on LIBERO-Plus and its strong performance on RoboTwin~2.0 and RoboDojo.
We summarize \textbf{our contributions} below.

\noindent \ding{182} \textbf{Supervision-wise}, to strengthen the supervision signal for higher performance, we propose visual residual learning. We show that this signal exhibits higher feature rank than full future features and concentrates on the robot and manipulated objects rather than static backgrounds, providing a more focused and expressive supervision signal.

\noindent \ding{183} \textbf{Architecture-wise}, to improve computational efficiency, we design \ours, a residual world-action model with 2.3B parameters in total, including a lightweight 0.5B dual-branch DiT that jointly denoises actions and frozen DINOv3 visual residuals via conditional flow matching in place of video generation models or separate world models.

\noindent \ding{184} \textbf{Experiment-wise}, without robot pretraining, \ours achieves the highest average success rates on LIBERO-Plus (81.1\%), RoboTwin~2.0 (93.8\%), and RoboDojo (9.40\% success, 13.41 score) at the lowest training (2.7--4.0\,TFLOPs) and inference compute (1.0--1.4\,TFLOPs), and leads real-world FR3 tasks (80.0\% average, $+10.0\%$ over $\pi_{0.5}$) at 92.7\,ms latency.

\section{Related Work}
\label{sec:related}

\paragraph{Generalist robot policies.}
Generalist policies combine visual-language priors with robot
demonstrations to transfer across tasks and embodiments. RT-2 and OpenVLA cast
discretized controls as language tokens~\cite{brohan2023rt,kim2024openvla}. For continuous
control, Octo uses diffusion over action chunks~\cite{ghosh2024octo}, $\pi_0$ and
$\pi_{0.5}$ introduce flow-matching action experts~\cite{black2024pi0,black2025pi0}, and
OpenVLA-OFT performs parallel action regression~\cite{fan2026rethinking,kim2025fine}. Video-based
pretraining further broadens supervision by extracting latent actions before
robot adaptation, as in LAPA and UniVLA~\cite{bu2025univla,ye2024latent}. These policies
predict only actions or latent actions, whereas \ours also predicts
visual residuals to supervise world change.

\paragraph{World models for robot control.}
Classical visual foresight rolls out action-conditioned video for goal
comparison~\cite{finn2017deep}. Video-generative WAMs produce visual
subgoals, interleave video and action generation, or jointly model video
latents and actions~\cite{bi2025motus,kim2026cosmos,li2026causal,lv2025f1,wang2026openwam}; Mantis
and Fast-WAM instead use visual or video prediction primarily during
training~\cite{yang2025mantis,yuan2026fast}. Feature-space methods predict latent futures:
DINO-WM and FLARE forecast or align future embeddings~\cite{zheng2025flare,zhou2024dino},
while others learn future
transitions, semantic or motion features, or latent subgoals~\cite{chen2026lawam,lin2026jepa,sun2026vla,won2025dual,yang2026lila,zhang2025dreamvla}. These methods
use complete future frames, video latents, or future features as their learning target, whereas
\ours replaces the complete-future learning target with a current-to-future visual
residual.

\paragraph{Temporal change modeling in world models.}
Recent methods model representations of temporal changes between current and future states
rather than the complete future. RLA-WM encodes frame-to-frame feature changes into
latent actions with an autoencoder and decodes them back into future
tokens~\cite{zhang2026learning}; Delta-JEPA decodes actions from latent
displacements~\cite{zhang2026delta}; and AD-WM regularizes latent state transitions through
action recovery for planning~\cite{qiu2026ad}. DeltaWAM jointly predicts visual
deltas and actions while retaining a dense anchor and a pretrained video
generator~\cite{yan2026deltawam}. However, these methods still require compressing temporal changes into latent actions and decoding them back to future features, recovering actions from latent displacements for planning, or coupling delta tokenizers with full-frame video generators;
\ours instead directly supervises frozen DINOv3 visual residuals alongside actions with a lightweight dual-branch DiT end to end.

\section{Preliminaries on VLA and WAM}
\label{sec:preliminaries}

\paragraph{Vision-language-action models.}
At time step $t$, the robot observes the camera images $\mathbf o_t$, its proprioceptive state $\mathbf q_t$, and a natural-language instruction $c$. Let $\hat{\mathbf a}_t\in\mathbb R^{d_a}$ denote a predicted robot action at step $t$. Instead of predicting a single action per query, the policy outputs an \emph{action chunk} $\hat\Act_t=\big[\hat{\mathbf a}_t,\hat{\mathbf a}_{t+1},\dots,\hat{\mathbf a}_{t+H-1}\big]^\top\in\mathbb R^{H\times d_a}$ containing the next $H$ predicted actions. The robot executes the predicted action chunk and then queries the policy again from the new observation. A VLA adapts a pretrained VLM to learn the conditional action distribution $\pi_{\bTheta}\big(\hat\Act_t\mid\mathbf o_t,\mathbf q_t,c\big)$, where
$\bTheta$ denotes all trainable policy parameters~\cite{black2024pi0,black2025pi0,kim2025fine}.

\paragraph{World-action models.}
Unlike VLAs, a WAM incorporates future world prediction into action learning, encouraging the model to anticipate the physical consequences of its actions. Let $\hat\Lat_{t+H}$ denote the predicted latent representation of the scene at time $t+H$, following execution of the predicted action chunk $\hat\Act_t$. A WAM jointly models actions and future world states through the conditional distribution $\pi_{\bTheta}\big(\hat\Act_t,\hat\Lat_{t+H}\mid\mathbf o_t,\mathbf q_t,c\big)$, coupling action generation with future visual prediction. A central design question is what \textit{supervision} target can most effectively capture action-induced visual changes without being dominated by static background content, which we investigate next.

\section{Rethinking World Supervision through Visual Residuals}
\label{sec:residual}

\subsection{From future states to visual residuals}
\label{sec:formulation}

Existing WAMs typically supervise world prediction by reconstructing complete future observations or latent representations. Specifically, given demonstration trajectories, a visual encoder $\Enc$ maps each observation into $N$ visual patch tokens, each of dimension $D$. Let $\Lat_t=\Enc(\mathbf o_t)\in\mathbb R^{N\times D}$ denote the current visual representation, where the $j$-th row $\mathbf z_{t,j}\in\mathbb R^D$ corresponds to patch $j$. Conventional WAMs use the full future representation as the supervision target for $\hat\Lat_{t+H}$, given by
\begin{equation}
  \Lat_{t+H} = \Enc(\mathbf o_{t+H})\in\mathbb R^{N\times D}.
  \label{eq:latents}%
\end{equation}%

However, over a relatively short horizon of $H$ steps, much of the scene remains unchanged, making the future representation $\Lat_{t+H}$ largely redundant with the current representation $\Lat_t$ and causing the future loss to be dominated by copying static background tokens. 
This motivates a natural question: \textit{Can we isolate and directly learn the changes between current and future observations, rather than predicting the entire future representation?} To this end, we define the \textbf{visual token residual} as the difference between future and current visual representations:
\begin{equation}
  \Res_t=\Lat_{t+H}-\Lat_t\in\mathbb R^{N\times D},
  \label{eq:residual}%
\end{equation}%
where each row $\res_{t,j}=\mathbf z_{t+H,j}-\mathbf z_{t,j}$ captures the feature change at patch $j$. From a saliency perspective, the residual suppresses redundant static scene content while emphasizing temporal changes, providing a more focused supervision signal for action-induced visual changes. 

Before incorporating $\Res_t$ into our policy architecture, we assess its suitability as a learning target from two complementary perspectives: (i) feature diversity, quantified by the rank of the feature matrix, which measures the number of linearly independent feature directions within a frame (\cref{sec:rank}); and (ii) spatial saliency, measured by the concentration of residual norms across image regions (\cref{sec:vis}).

\subsection{The residual has higher rank than the future}
\label{sec:rank}

To assess the suitability of the visual residual $\Res_t$ vs.\ the future representation $\Lat_{t+H}$ as a learning target, we compare their \textit{effective rank} and \textit{stable rank}, which characterize the diversity of feature directions within a single frame. For $\mathbf X\in\{\Lat_{t+H},\Res_t\}$, we first center its visual tokens using $\mathbf J\mathbf X$, where $\mathbf J=\mathbf I_N-\frac{1}{N}\mathbf 1\mathbf 1^\top$. Let $\{ \sigma_i\}$ denote the singular values of the centered feature matrix, and define the normalized spectral energy distribution as $p_i=\sigma_i^2/\sum_{i'}\sigma_{i'}^2$. The effective rank and stable rank are then defined as
\begin{equation}
  \reff(\mathbf X)=\exp\big(\mathcal H(\mathbf p)\big), \quad 
  \rsta(\mathbf X)=\frac{\sum_i\sigma_i^2}{\sigma_1^2},
  \label{eq:ranks}%
\end{equation}%
where $\mathcal H(\mathbf p)=-\sum_ip_i\log p_i$ is the Shannon entropy.
Here $\reff$ measures the effective number of directions that carry energy, and
$\rsta$ is the inverse energy share of the dominant direction. Both metrics are scale-invariant, allowing a direct comparison of the spectral shapes even though the residual has smaller magnitude.

\begin{figure}[!htb]
  \centering
  \includegraphics[width=0.60\linewidth]{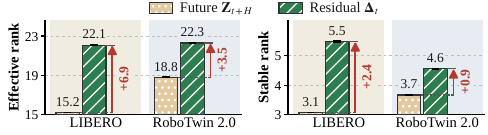}
  \caption{Effective rank (left) and stable rank (right) of the future DINOv3
  tokens $\mathbf Z_{t+H}$ and the residual $\boldsymbol\Delta_t$, averaged over
  all windows of LIBERO and RoboTwin~2.0.}
  \label{fig:rank}
\end{figure}

\textbf{\cref{fig:rank}} shows the effective rank and stable rank of the future tokens $\Lat_{t+H}$ and the residual $\Res_t$ with DINOv3~\cite{simeoni2025dinov3} across all windows of LIBERO~\cite{liu2023libero} and RoboTwin~2.0~\cite{chen2025robotwin} (see \textbf{\cref{app:rank}} for experimental details). As we can see, the residual consistently spans more feature directions than the future on both benchmarks: averaged over all windows, the effective rank rises from 15.22 for the future to 22.06 for the residual on LIBERO ($+6.84$) and from 18.79 to 22.28 on RoboTwin~2.0 ($+3.49$), while the stable rank rises from 3.08 to 5.48 ($+2.40$) and from 3.67 to 4.56 ($+0.89$). This indicates that learning the residual does not simplify the learning target; rather, it changes what the learning target emphasizes. Supervising the complete future is heavily weighted toward reproducing a few dominant modes of static background content, whereas the residual removes this unchanged component and distributes supervision across more, individually weaker directions of change, providing the residual branch with a richer and more diverse supervisory signal for robot and object motion.

\subsection{The residual can concentrate on the robot}
\label{sec:vis}

\begin{figure}[!htb]
  \centering
  \small
  \setlength{\tabcolsep}{0.6pt}
  \renewcommand{\arraystretch}{0}
  \newcommand{\vistile}[1]{\raisebox{-0.5\height}{#1}}
  \newcommand{\vislabel}[1]{\rotatebox[origin=c]{90}{#1}}
  \begin{tabular}{@{}c@{\hspace{2pt}}cccc@{}}
    & $\mathbf o_t$ & $\mathbf o_{t+H}$ & Saliency map & PCA-RGB \\ \noalign{\vskip 4pt}
    \vislabel{LIBERO}
      & \vistile{\includegraphics[width=0.13\linewidth]{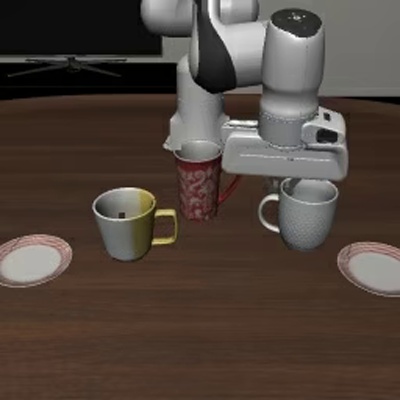}}
      & \vistile{\includegraphics[width=0.13\linewidth]{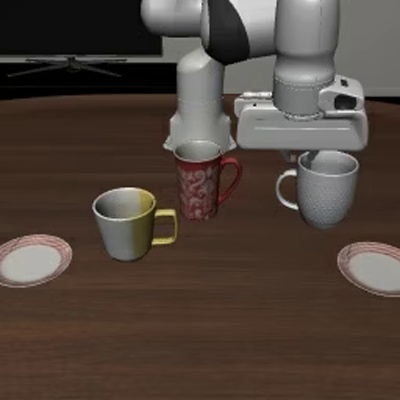}}
      & \vistile{\includegraphics[width=0.13\linewidth]{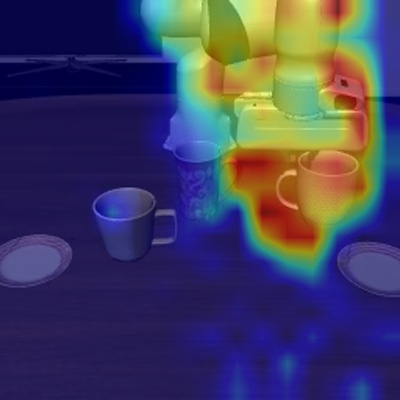}}
      & \vistile{\includegraphics[width=0.13\linewidth]{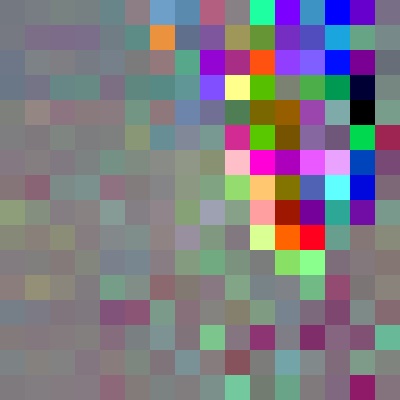}} \\ \noalign{\vskip 1.2pt}
    \vislabel{RoboTwin~2.0}
      & \vistile{\includegraphics[width=0.13\linewidth]{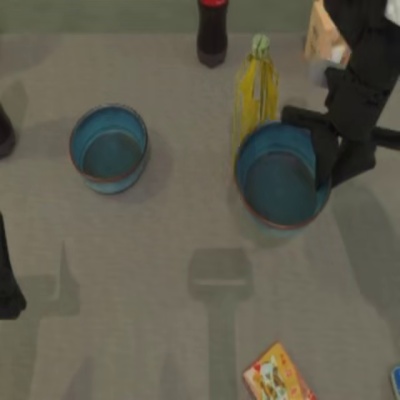}}
      & \vistile{\includegraphics[width=0.13\linewidth]{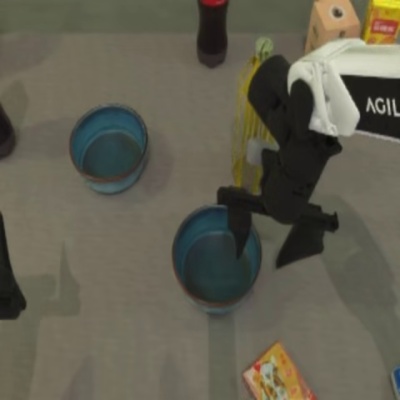}}
      & \vistile{\includegraphics[width=0.13\linewidth]{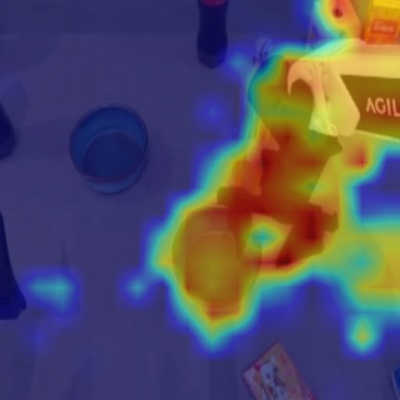}}
      & \vistile{\includegraphics[width=0.13\linewidth]{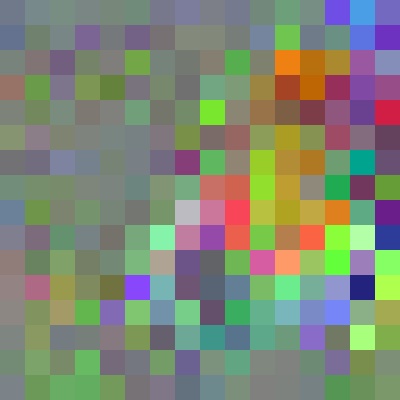}} \\
  \end{tabular}
  \caption{Residual projected back onto the
  observation. Saliency map: per-token residual norm
  overlaid on $\mathbf o_{t+H}$. PCA-RGB: residual tokens projected onto three fixed principal directions.}
  \label{fig:vis}
\end{figure}

Next, we visualize the residual tokens in image space using two complementary approaches (\textbf{\cref{fig:vis}}): (i) a \emph{saliency map}, which overlays the per-token residual magnitude $\|\res_{t,j}\|_2$ on the future frame to highlight regions of significant change, and (ii) a \emph{PCA-RGB map}, which projects residual tokens onto the top three principal components and maps them to RGB channels to reveal their spatial structure (\textbf{\cref{app:vis}}).

As we can see from \cref{fig:vis}, the strongest residual responses closely follow the moving robot arm, the end effector, and the manipulated object, while the most saturated PCA colors appear along the trajectory swept by the arm; on RoboTwin~2.0, the residual also clearly marks both the initial and the displaced positions of the carried bowl, whereas the surrounding background remains nearly inactive. These observations confirm that subtracting the current tokens naturally filters out static scene layout and concentrates supervision on the gripper and the manipulated objects, providing the world-action model with a spatially focused learning signal aligned with physical interaction.

\section{ResWAM: Residual World-Action Model}
\label{sec:method}

\subsection{Model architecture}
\label{sec:arch}

\begin{figure}[!htb]
  \centering
  \includegraphics[width=0.6\linewidth]{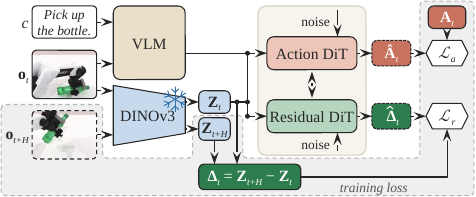}
  \caption{\ours architecture overview. 
  A VLM encodes the current images $\mathbf o_t$ and
  instruction $c$. The action and residual
  branches denoise the predicted action chunk $\hat\Act_t$ and residual
  $\hat\Res_t$ by cross-attending to the VLM output and the current DINOv3 tokens
  $\Lat_t$, exchanging information via joint attention. Here $\mathcal L_a$
  and $\mathcal L_r$ are the flow-matching losses supervised by $\Act_t$ and
  $\Res_t=\Lat_{t+H}-\Lat_t$, and \ding{100} marks the frozen encoder.}
  \label{fig:arch}
\end{figure}

Motivated by the properties of the visual token residual established in \textbf{\cref{sec:residual}}, we formulate \ours by replacing the complete future learning target of conventional WAMs (as introduced in \textbf{\cref{sec:preliminaries}} and compared in \cref{fig:teaser}) 
with the residual learning target (\textbf{\cref{fig:arch}}). Specifically, conditioned on the current camera images $\mathbf o_t$, proprioceptive state $\mathbf q_t$, and language instruction $c$, \ours models the joint distribution of the predicted action chunk $\hat\Act_t\in\mathbb R^{H\times d_a}$ and the predicted visual token residual $\hat\Res_t\in\mathbb R^{N\times D}$:
\begin{equation}
  \pi_{\bTheta}\big(\hat\Act_t,\hat\Res_t\mid\mathbf o_t,\mathbf q_t,c\big).
  \label{eq:reswam}%
\end{equation}%

We adopt RynnBrain-1.1-2B~\cite{li2026rynnbrain} as the VLM backbone, which jointly encodes all current camera views and language instructions into output tokens, with all backbone parameters trained end-to-end. Building on the VLM output, we introduce two coupled branches: an \emph{action branch} (\textit{i.e.}, ``Action DiT'' in \cref{fig:arch}) predicting action chunks $\hat\Act_t$ and a \emph{residual branch} (\textit{i.e.}, ``Residual DiT'' in \cref{fig:arch}) predicting visual token residuals $\hat\Res_t$, each from its respective noise. 
Both branches employ DiT-style~\cite{peebles2023scalable} joint transformer blocks with branch-specific parameters and flow-time conditioning, while bidirectional attention over concatenated action and residual tokens enables information exchange. Each branch further cross-attends to the VLM output and current visual tokens $\Lat_t\in\mathbb R^{N\times D}$ extracted by a frozen DINOv3~\cite{simeoni2025dinov3} encoder ($\Enc$ in \textbf{\cref{eq:latents}}), where $2\times2$ spatial average pooling compresses the $16\times16$ patch grid into $N=64$ tokens per view.

Unlike conventional WAM architectures (\cref{fig:teaser}(b)) that either rely on large video generation backbones (such as Wan~\cite{bi2025motus,li2026causal,yuan2026fast} and Cosmos~\cite{kim2026cosmos}) and 3D video VAEs to reconstruct future pixels or maintain a separate world model alongside a VLA~\cite{chen2026lawam}, \ours attaches a lightweight 0.5B dual-branch DiT directly to a pretrained VLM and a frozen DINOv3 encoder, totaling only 2.3B parameters and keeping both training and inference compute lightweight.

\subsection{Training and inference mechanisms}
\label{sec:objective}

\paragraph{Training mechanism.}
Both branches are trained with conditional flow matching~\cite{lipman2023flow}.
For each branch $s\in\{a,r\}$, we flatten the learning target into a vector
$\mathbf y^s$, sample noise $\bm\epsilon^s\sim\mathcal N(\mathbf 0,\mathbf I)$
and a flow time $\tau_s\in[0,1]$, form
$\mathbf x^s_{\tau_s}=(1-\tau_s)\bm\epsilon^s+\tau_s\mathbf y^s$, and regress
the velocity $\mathbf u^s=\mathbf y^s-\bm\epsilon^s$. The action learning target is
the normalized action chunk $\Act_t$ from the demonstration. The world learning target
is the residual of \textbf{\cref{eq:residual}} between the DINOv3
tokens of the future and current frames (\cref{fig:arch}), computed on the fly by the frozen encoder from the frames at $t$ and $t+H$ of
the same demonstration. The two
branches draw independent noise and flow times, and the loss is
\begin{equation}
  \begin{aligned}
  &\mathcal L=\mathcal L_a+\lambda_w\mathcal L_r, ~~\text{where}~~\\
 & \mathcal L_a=\mathbb E\,\big\|\mathbf v_{\bTheta}^{a}-\mathbf u^a\big\|_2^2,\quad
  \mathcal L_r=\mathbb E\,\big\|\mathbf v_{\bTheta}^{w}-\mathbf u^w\big\|_2^2.
  \end{aligned}
  \label{eq:loss}%
\end{equation}%
In \textbf{\cref{eq:loss}}, $\mathbf v_{\bTheta}^a$ and $\mathbf v_{\bTheta}^w$ are the action and world
velocities predicted from both noisy branches
$(\mathbf x^a_{\tau_a},\mathbf x^w_{\tau_w})$, both flow times, and the inputs
$(\mathbf o_t,\mathbf q_t,c)$ of \textbf{\cref{eq:reswam}}, and $\mathbf u^a$ and
$\mathbf u^w$ are their target velocities. Because the residual focuses on the robot and manipulated objects, even without a large pretrained video generation model or separate world model, \ours obtains effective supervision from the residual branch to help the action branch attend to task-relevant regions and anticipate the consequences of its actions.

\paragraph{Inference mechanism.}
At test time, both the action and residual branches start from Gaussian noise and are integrated jointly over 10 Euler steps (\cref{fig:arch}). Unlike video-generation WAMs that denoise high-dimensional video latents and decode future pixels through heavy 3D VAEs~\cite{bi2025motus,kim2026cosmos,li2026causal}, or external-world-model WAMs that query a separate world model before action generation~\cite{chen2026lawam}, \ours denoises compact residual tokens in latent space alongside the action chunk in a single forward pass without pixel-space decoding, enabling low-latency real-time control.

\section{Experiments}
\label{sec:experiments}

\subsection{Experimental setup}
\label{sec:setup}

\paragraph{Model and training setup.}
\ours combines RynnBrain-1.1-2B~\cite{li2026rynnbrain} as the VLM backbone, a frozen DINOv3 ViT-B/16~\cite{simeoni2025dinov3} encoder ($2\times2$-pooled to $N=64$ tokens), and a 0.5B dual-branch DiT~\cite{peebles2023scalable}, trained directly on downstream demonstrations with AdamW (\textbf{\cref{app:training}}) without robot pretraining.

\begin{figure}[!htb]
  \centering
  \includegraphics[width=0.72\linewidth]{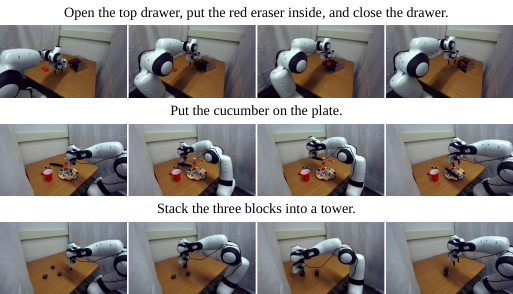}
  \caption{Illustrations of the three real-world manipulation tasks.}
  \label{fig:real-robot-main}
\end{figure}

\paragraph{Benchmarks and evaluation protocols.}
We train and evaluate one multi-task policy per benchmark:
(1)~on \textbf{LIBERO}~\cite{liu2023libero}, we train across all four standard suites (\emph{Spatial}, \emph{Object}, \emph{Goal}, \emph{Long}) and evaluate across all 10{,}030 perturbed episodes of \textbf{LIBERO-Plus}~\cite{fei2025libero} without exposure to perturbations during training;
(2)~on \textbf{RoboTwin~2.0}~\cite{chen2025robotwin}, we train on clean and domain-randomized demonstrations across all 50 bimanual tasks and evaluate 100 trials per task under both clean and randomized conditions;
(3)~on \textbf{RoboDojo}~\cite{chen2026robodojo}, we train on the 35-task training set and evaluate across all 42 tasks (2{,}100 episodes) spanning five capability dimensions; and
(4)~in \textbf{real-world} manipulation, we train and evaluate over 30 trials per task on a Franka Research 3 (FR3) arm following DROID~\cite{khazatsky2024droid} across three tasks (\textbf{\cref{fig:real-robot-main}}): \emph{Open} (opening the top drawer, placing a red eraser inside, and closing it) for multi-stage articulated manipulation, \emph{Put} (placing a cucumber onto a plate) for elongated-object grasping and placement, and \emph{Stack} (stacking three blocks) for sub-centimeter alignment.

\paragraph{Metrics and baselines.}
For task performance, we report task \emph{success rate} (\%) (plus stage-level progress \emph{Score} on RoboDojo); for computational efficiency, we report per-sample training compute (\emph{Train}, TFLOPs) and per-chunk inference compute (\emph{Infer.}, TFLOPs) at batch size 1, alongside per-chunk inference \emph{Latency} (ms) averaged over 1{,}000 iterations on one NVIDIA RTX~A6000 (\textbf{\cref{app:latency}}; standard deviations in \textbf{\cref{fig:latency-std}}).
We compare \ours against representative \textbf{VLAs} (OpenVLA~\cite{kim2024openvla}, Mantis~\cite{yang2025mantis}, UniVLA~\cite{bu2025univla}, $\pi_0$~\cite{black2024pi0}, $\pi_{0.5}$~\cite{black2025pi0}, OpenVLA-OFT~\cite{kim2025fine}, GR00T-N1.6~\cite{bjorck2025gr00t}, VLA-JEPA~\cite{sun2026vla}, InternVLA-A1~\cite{cai2026internvla}, LingBot-VLA~\cite{wu2026pragmatic}, and StarVLA-$\alpha$~\cite{ye2026starvla}) and \textbf{WAMs} (Motus~\cite{bi2025motus}, LingBot-VA~\cite{li2026causal}, LaWAM~\cite{chen2026lawam}, Fast-WAM~\cite{yuan2026fast}, Cosmos-Policy~\cite{kim2026cosmos}, LDA-1B~\cite{lyu2026lda}, AHA-WAM~\cite{cai2026aha}, and GigaWorld-Policy~\cite{ye2026gigaworld}).

\subsection{Experimental results}
\label{sec:sim}

\begin{table}[!htb]
  \centering
  \caption{Efficiency and success rates (\%) on LIBERO-Plus across the seven perturbation dimensions and their overall average (Avg.) over all 10{,}030 perturbed episodes.
  All policies are trained on standard LIBERO demonstrations (without exposure to LIBERO-Plus perturbations).
  \textit{Robot Pretrain} indicates whether a policy uses external robot action datasets beyond the benchmark demonstrations (\cmark) or is trained solely on benchmark demonstrations (\xmark).
  \textit{Size} is the number of parameters in billions loaded at inference (\textbf{\cref{app:params}}).
  \textit{Efficiency} comprises per-sample training (\textit{Train}) and per-chunk inference (\textit{Infer.}) compute in TFLOPs, and per-chunk inference \textit{Latency} in ms on one NVIDIA RTX A6000 (\cref{app:latency}).
  The \textbf{best} result in each column is in \textbf{bold} and the \underline{second best} is \underline{underlined}.}
  \label{tab:libero-plus}
  \scriptsize
  \setlength{\arrayrulewidth}{0.5pt}
  \setlength{\tabcolsep}{3.8pt}
  \begin{tabular}{@{}l|>{\vr}cc|>{\vr}ccc|>{\vr}ccccccc!{\dashedvrule}>{\dvr}c@{}}
    \toprule[1pt]
    \multirow{2}{*}[-0.2ex]{Method} & Robot & \multirow{2}{*}[-0.2ex]{Size} & \multicolumn{3}{c|}{Efficiency} & \multicolumn{8}{c@{}}{Success Rate (\%)} \\
    \cline{4-14}
    \rule{0pt}{2.2ex} & Pretrain & & Train & Infer. & Latency & Camera & Robot & Language & Light & Background & Noise & Layout & Avg. \\
    \midrule
    \rowcolor{vlarow}\multicolumn{14}{@{}l@{}}{\textit{Vision-language-action models (VLAs)}} \\
    \rowcolor{vlarow}OpenVLA~\cite{kim2024openvla} & \cmark & 7.5 & 13.0 & 5.0 & 235.1 & 0.8 & 3.5 & 23.0 & 8.1 & 34.8 & 15.2 & 28.5 & 15.6 \\
    \rowcolor{vlarow}Mantis~\cite{yang2025mantis} & \cmark & 5.8 & 19.3 & 5.6 & 311.1 & 39.2 & 15.7 & 41.8 & 62.5 & 28.9 & 45.9 & 45.1 & 39.8 \\
    \rowcolor{vlarow}UniVLA~\cite{bu2025univla} & \cmark & 7.6 & 9.4 & 4.5 & 166.3 & 1.8 & 46.2 & 69.6 & 69.0 & 81.0 & 21.2 & 31.9 & 42.9 \\
    \rowcolor{vlarow}$\pi_0$~\cite{black2024pi0} & \cmark & 3.2 & 11.8 & 4.4 & \second{112.2} & 13.8 & 6.0 & 58.8 & 85.0 & 81.4 & 79.0 & 68.9 & 53.6 \\
    \rowcolor{vlarow}$\pi_{0.5}$~\cite{black2025pi0} & \cmark & 3.4 & 13.5 & 4.7 & 120.1 & 43.2 & 49.4 & 71.1 & 92.8 & 86.2 & 48.0 & 79.4 & 65.8 \\
    \rowcolor{vlarow}OpenVLA-OFT~\cite{kim2025fine} & \cmark & 7.7 & 18.4 & 8.9 & 132.4 & 56.4 & 31.9 & 79.5 & 88.7 & 93.3 & 75.8 & 74.2 & 69.6 \\
    \rowcolor{vlarow}GR00T-N1.6~\cite{bjorck2025gr00t} & \cmark & 3.3 & \second{3.2} & 2.3 & 244.9 & \second{64.4} & 38.4 & 83.9 & 94.9 & 93.7 & \second{84.4} & 74.9 & 74.9 \\
    \rowcolor{vlarow}VLA-JEPA~\cite{sun2026vla} & \cmark & 2.8 & 5.7 & \second{1.2} & 192.3 & 63.3 & 67.1 & 85.4 & \second{95.6} & 93.6 & 66.3 & \best{85.1} & \second{79.5} \\
    \midrule
    \rowcolor{wamrow}\multicolumn{14}{@{}l@{}}{\textit{World-action models (WAMs)}} \\
    \rowcolor{wamrow}Motus~\cite{bi2025motus} & \cmark & 14.4 & 16.0 & 76.1 & 2769.5 & 17.9 & 43.3 & 74.1 & 54.7 & 44.9 & 26.2 & 52.0 & 44.0 \\
    \rowcolor{wamrow}LingBot-VA~\cite{li2026causal} & \cmark & 11.5 & 14.2 & 64.6 & 2903.7 & 40.9 & \best{83.0} & 86.4 & 82.3 & 53.1 & 64.4 & 76.2 & 69.5 \\
    \rowcolor{wamrow}LaWAM~\cite{chen2026lawam} & \cmark & 2.6 & \second{3.2} & \second{1.2} & 198.1 & 37.2 & \second{71.6} & \second{87.1} & 93.5 & \second{94.1} & 69.4 & \second{83.3} & 76.6 \\
    \rowcolor{wamrow}Fast-WAM~\cite{yuan2026fast} & \xmark & 12.4 & 12.0 & 3.7 & 142.8 & 16.4 & 44.5 & 68.9 & 78.2 & 53.7 & 37.7 & 60.7 & 51.5 \\
    \rowcolor{wamrow}Cosmos-Policy~\cite{kim2026cosmos} & \xmark & 2.1 & 29.3 & 76.8 & 1123.9 & \best{69.7} & 42.4 & 86.3 & 90.9 & 77.5 & 75.6 & 65.2 & 71.5 \\
    \midrule
    \rowcolor{oursrow}\textbf{\ours (Ours)} & \xmark & 2.3 & \best{2.7} & \best{1.0} & \best{92.2} & 52.3 & 71.0 & \best{89.9} & \best{97.7} & \best{98.3} & \best{85.6} & 83.2 & \best{81.1} \\
    \bottomrule[1pt]
  \end{tabular}
\end{table}

\paragraph{LIBERO-Plus.}
\label{sec:libero-plus}
To assess out-of-distribution generalization and computational efficiency, \textbf{\cref{tab:libero-plus}} evaluates policies trained only on standard LIBERO demonstrations across all 10{,}030 perturbed episodes of LIBERO-Plus~\cite{fei2025libero}. Without robot pretraining and with 2.3B parameters, \ours achieves an 81.1\% overall success rate, ranking first among all 14 compared methods and outperforming the strongest VLA baseline, VLA-JEPA (79.5\%), and the strongest WAM baseline, LaWAM (76.6\%), by $+1.6$ and $+4.5$ points, respectively. Notably, pixel-level video WAMs such as Fast-WAM (51.5\%, 12.4B) and Motus (44.0\%, 14.4B) degrade sharply under Background (53.7\% and 44.9\%) and Sensor Noise (37.7\% and 26.2\%) shifts because reconstructing complete future frames couples policy representations to background appearance and pixel statistics. By supervising only the DINOv3 visual token residual, \ours achieves the highest success rates on Language (89.9\%), Light (97.7\%), Background (98.3\%), and Noise (85.6\%).
In the Efficiency columns of \cref{tab:libero-plus}, \ours requires the lowest training compute (2.7\,TFLOPs), inference compute (1.0\,TFLOPs, $3.7\times$--$4.7\times$ below Fast-WAM and $\pi_{0.5}$, and $65\times$--$77\times$ below LingBot-VA, Motus, and Cosmos-Policy), and latency (92.2\,ms), running $2.1\times$ faster than LaWAM (198.1\,ms) and over an order of magnitude faster than multi-step video WAMs.

\begin{table}[!htb]
  \centering
  \caption{Efficiency and success rates (\%) on RoboTwin~2.0 (50 tasks, 100 trials per task) under clean and randomized (Random) conditions. Notation as in \cref{tab:libero-plus}.}
  \label{tab:robotwin}
  \footnotesize
  \setlength{\arrayrulewidth}{0.5pt}
  \setlength{\tabcolsep}{3.0pt}
  \begin{tabular}{@{}l|>{\vr}cc|>{\vr}ccc|>{\vr}cc!{\dashedvrule}>{\dvr}c@{}}
    \toprule[1pt]
    \multirow{2}{*}[-0.2ex]{Method} & Robot & \multirow{2}{*}[-0.2ex]{Size} & \multicolumn{3}{c|}{Efficiency} & \multicolumn{3}{c@{}}{Success Rate (\%)} \\
    \cline{4-9}
    \rule{0pt}{2.2ex} & Pretrain & & Train & Infer. & Latency & Clean & Random & Avg. \\
    \midrule
    \rowcolor{vlarow}\multicolumn{9}{@{}l@{}}{\textit{Vision-language-action models (VLAs)}} \\
    \rowcolor{vlarow}$\pi_0$~\cite{black2024pi0} & \cmark & 3.2 & 11.8 & 4.4 & \second{111.9} & 65.9 & 58.4 & 62.2 \\
    \rowcolor{vlarow}$\pi_{0.5}$~\cite{black2025pi0} & \cmark & 3.4 & 13.6 & 5.0 & 124.7 & 82.7 & 76.8 & 79.8 \\
    \midrule
    \rowcolor{wamrow}\multicolumn{9}{@{}l@{}}{\textit{World-action models (WAMs)}} \\
    \rowcolor{wamrow}Motus~\cite{bi2025motus} & \cmark & 14.4 & 16.1 & 76.6 & 2902.5 & 88.7 & 87.0 & 87.8 \\
    \rowcolor{wamrow}LaWAM~\cite{chen2026lawam} & \cmark & 2.6 & \second{4.1} & \second{1.5} & 220.9 & 92.6 & 89.8 & 91.2 \\
    \rowcolor{wamrow}LingBot-VA~\cite{li2026causal} & \cmark & 11.5 & 23.1 & 114.7 & 4473.7 & \second{92.9} & 91.6 & \second{92.2} \\
    \rowcolor{wamrow}Fast-WAM~\cite{yuan2026fast} & \xmark & 12.4 & 14.6 & 4.0 & 148.8 & 91.9 & \second{91.8} & 91.8 \\
    \midrule
    \rowcolor{oursrow}\textbf{\ours (Ours)} & \xmark & 2.3 & \best{4.0} & \best{1.4} & \best{99.4} & \best{93.6} & \best{94.0} & \best{93.8} \\
    \bottomrule[1pt]
  \end{tabular}
\end{table}

\paragraph{RoboTwin 2.0.}
\label{sec:robotwin}
\textbf{\cref{tab:robotwin}} reports the evaluation results across all 50 bimanual tasks of RoboTwin~2.0. \ours achieves 93.6\% under clean conditions and 94.0\% under domain-randomized conditions, with an overall average of 93.8\% at the lowest training compute (4.0\,TFLOPs), inference compute (1.4\,TFLOPs), and latency (99.4\,ms), ranking first in every column of \cref{tab:robotwin}. Without robot pretraining, \ours outperforms the VLAs $\pi_0$ (62.2\%, 4.4\,TFLOPs, 111.9\,ms) and $\pi_{0.5}$ (79.8\%, 5.0\,TFLOPs, 124.7\,ms) by $+31.6$ and $+14.0$ points. Among WAMs, \ours surpasses Fast-WAM (91.8\%, 14.6\,TFLOPs train, 148.8\,ms) and LingBot-VA (92.2\%, 114.7\,TFLOPs infer., 4{,}473.7\,ms) by $+2.0$ and $+1.6$ points with $3.6\times$--$5.8\times$ lower training compute, and beats LaWAM (91.2\%, 220.9\,ms) and Motus (87.8\%, 2{,}902.5\,ms) by $+2.6$ and $+6.0$ points.

\begin{table}[!htb]
  \centering
  \caption{Efficiency, success rates (SR, \%), and stage-level progress scores (Score) on the RoboDojo simulation benchmark across the five capability dimensions (42 tasks, 54 configurations, 2{,}100 episodes) and their macro-average (Avg.). Notation as in \cref{tab:libero-plus}.}
  \label{tab:robodojo}
  \scriptsize
  \setlength{\arrayrulewidth}{0.5pt}
  \setlength{\tabcolsep}{2.7pt}
  \begin{tabular}{@{}l|>{\vr}cc|>{\vr}ccc|>{\vr}cc|>{\vr}cc|>{\vr}cc|>{\vr}cc|>{\vr}cc!{\dashedvrule}>{\dvr}cc@{}}
    \toprule[1pt]
    \multirow{2}{*}[-0.2ex]{Method} & Robot & \multirow{2}{*}[-0.2ex]{Size} & \multicolumn{3}{c|}{Efficiency} & \multicolumn{2}{c|}{Generalization} & \multicolumn{2}{c|}{Precision} & \multicolumn{2}{c|}{Long-Horizon} & \multicolumn{2}{c|}{Memory} & \multicolumn{2}{c!{\dashedvrule}}{Open} & \multicolumn{2}{c@{}}{Avg.} \\
    \cline{4-18}
    \rule{0pt}{2.2ex} & Pretrain & & Train & Infer. & Latency & SR & Score & SR & Score & SR & Score & SR & Score & SR & Score & SR & Score \\
    \midrule
    \rowcolor{vlarow}\multicolumn{18}{@{}l@{}}{\textit{Vision-language-action models (VLAs)}} \\
    \rowcolor{vlarow}OpenVLA-OFT~\cite{kim2025fine} & \cmark & 7.7 & 35.7 & 17.2 & 237.7 & 0.00 & 0.04 & 0.00 & 0.20 & 0.00 & 0.70 & 0.00 & 0.00 & 0.08 & 0.08 & 0.02 & 0.21 \\
    \rowcolor{vlarow}InternVLA-A1~\cite{cai2026internvla} & \cmark & 3.2 & 9.0 & \second{2.2} & 422.5 & 1.83 & 2.87 & 0.92 & 3.00 & 1.17 & 4.79 & 1.33 & 1.58 & 0.17 & 0.17 & 1.08 & 2.48 \\
    \rowcolor{vlarow}GR00T-N1.6~\cite{bjorck2025gr00t} & \cmark & 3.3 & 6.1 & 3.6 & 259.8 & 1.22 & 2.16 & 0.67 & 2.54 & 3.58 & 8.30 & 0.89 & 1.06 & 0.17 & 0.18 & 1.31 & 2.85 \\
    \rowcolor{vlarow}$\pi_0$~\cite{black2024pi0} & \cmark & 3.2 & 11.8 & 4.4 & \second{112.5} & 2.56 & 3.94 & 0.75 & 3.56 & 2.00 & 6.19 & 2.11 & 3.47 & 0.25 & 0.25 & 1.53 & 3.48 \\
    \rowcolor{vlarow}LingBot-VLA~\cite{wu2026pragmatic} & \cmark & 4.2 & 6.7 & 2.6 & 740.4 & 4.28 & 6.71 & 1.83 & 5.33 & 5.25 & 10.89 & 2.78 & 3.82 & 0.67 & 0.72 & 2.96 & 5.50 \\
    \rowcolor{vlarow}$\pi_{0.5}$~\cite{black2025pi0} & \cmark & 3.4 & 13.6 & 5.0 & 125.4 & \best{8.17} & \best{13.37} & \second{5.50} & \second{12.40} & \second{14.67} & \second{23.54} & \second{4.56} & \second{5.78} & \best{1.67} & \best{1.98} & \second{6.91} & \second{11.41} \\
    \rowcolor{vlarow}StarVLA-$\alpha$~\cite{ye2026starvla} & \xmark & 4.6 & 7.0 & 2.6 & 192.8 & 2.33 & 3.93 & 4.33 & 9.90 & 6.50 & 14.15 & 2.44 & 3.34 & 0.58 & 0.68 & 3.24 & 6.40 \\
    \midrule
    \rowcolor{wamrow}\multicolumn{18}{@{}l@{}}{\textit{World-action models (WAMs)}} \\
    \rowcolor{wamrow}LDA-1B~\cite{lyu2026lda} & \cmark & 5.4 & \second{5.9} & 3.7 & 358.2 & 0.17 & 0.71 & 0.50 & 3.21 & 0.08 & 1.92 & 1.78 & 2.08 & 0.00 & 0.00 & 0.51 & 1.58 \\
    \rowcolor{wamrow}Fast-WAM~\cite{yuan2026fast} & \xmark & 12.4 & 14.6 & 4.0 & 148.3 & 1.11 & 2.34 & 0.00 & 1.96 & 5.17 & 9.14 & 3.44 & 3.55 & 0.42 & 0.42 & 2.03 & 3.48 \\
    \rowcolor{wamrow}AHA-WAM~\cite{cai2026aha} & \xmark & 13.7 & 33.8 & 3.2 & 637.0 & 3.28 & 5.79 & 2.42 & 5.86 & 2.67 & 8.61 & 2.78 & 2.97 & 0.83 & 0.88 & 2.39 & 4.82 \\
    \rowcolor{wamrow}GigaWorld-Policy~\cite{ye2026gigaworld} & \xmark & 12.4 & 34.6 & 2.5 & 666.3 & 2.89 & 5.34 & 1.83 & 6.15 & 8.92 & 15.51 & 2.22 & 3.46 & 0.50 & 0.54 & 3.27 & 6.20 \\
    \midrule
    \rowcolor{oursrow}\textbf{\ours (Ours)} & \xmark & 2.3 & \best{3.9} & \best{1.4} & \best{99.7} & \second{6.50} & \second{9.12} & \best{14.75} & \best{21.31} & \best{15.75} & \best{25.52} & \best{9.00} & \best{9.97} & \second{1.00} & \second{1.14} & \best{9.40} & \best{13.41} \\
    \bottomrule[1pt]
  \end{tabular}
\end{table}

\paragraph{RoboDojo.}
\label{sec:robodojo}
\textbf{\cref{tab:robodojo}} evaluates \ours on the RoboDojo bimanual benchmark~\cite{chen2026robodojo} across five capability dimensions. Without robot pretraining, \ours achieves the highest overall success rate (9.40\%) and progress score (13.41) alongside the lowest training compute (3.9\,TFLOPs), inference compute (1.4\,TFLOPs), and latency (99.7\,ms) among all compared VLAs and WAMs, outperforming $\pi_{0.5}$ (6.91\% SR, 11.41 Score, 5.0\,TFLOPs, 125.4\,ms) by $+2.49$ points in SR and $+2.00$ points in Score, and nearly tripling GigaWorld-Policy (3.27\% SR, 6.20 Score, 34.6\,TFLOPs train, 666.3\,ms) at $6.7\times$ lower latency and $8.9\times$ lower training compute. In particular, \ours leads by a wide margin on Precision (14.75\% vs.\ 5.50\% for $\pi_{0.5}$), Long-Horizon (15.75\% vs.\ 14.67\%), and Memory (9.00\% vs.\ 4.56\%), showing that supervising the policy with DINOv3 token residuals strengthens contact-rich alignment and multi-stage bimanual execution.

\begin{table}[!htb]
  \centering
  \caption{Efficiency and success rates (\%) on the three real-world manipulation tasks (30 trials per task). Notation as in \cref{tab:libero-plus}.}
  \label{tab:real}
  \footnotesize
  \setlength{\arrayrulewidth}{0.5pt}
  \setlength{\tabcolsep}{3.0pt}
  \begin{tabular}{@{}l|>{\vr}cc|>{\vr}ccc|>{\vr}ccc!{\dashedvrule}>{\dvr}c@{}}
    \toprule[1pt]
    \multirow{2}{*}[-0.2ex]{Method} & Robot & \multirow{2}{*}[-0.2ex]{Size} & \multicolumn{3}{c|}{Efficiency} & \multicolumn{4}{c@{}}{Success Rate (\%)} \\
    \cline{4-10}
    \rule{0pt}{2.2ex} & Pretrain & & Train & Infer. & Latency & Open & Put & Stack & Avg. \\
    \midrule
    \rowcolor{vlarow}\multicolumn{10}{@{}l@{}}{\textit{Vision-language-action models (VLAs)}} \\
    \rowcolor{vlarow}GR00T-N1.6~\cite{bjorck2025gr00t} & \cmark & 3.3 & \second{3.2} & 2.3 & 245.5 & 60.0 & 76.7 & 23.3 & 53.3 \\
    \rowcolor{vlarow}$\pi_{0.5}$~\cite{black2025pi0} & \cmark & 3.4 & 13.6 & 5.0 & \second{120.4} & \second{80.0} & 86.7 & \second{43.3} & \second{70.0} \\
    \midrule
    \rowcolor{wamrow}\multicolumn{10}{@{}l@{}}{\textit{World-action models (WAMs)}} \\
    \rowcolor{wamrow}LaWAM~\cite{chen2026lawam} & \cmark & 2.6 & \second{3.2} & \second{1.2} & 198.8 & \second{80.0} & \second{90.0} & 36.7 & 68.9 \\
    \rowcolor{wamrow}Fast-WAM~\cite{yuan2026fast} & \xmark & 12.4 & 12.0 & 3.7 & 143.6 & 56.7 & 63.3 & 30.0 & 50.0 \\
    \midrule
    \rowcolor{oursrow}\textbf{\ours (Ours)} & \xmark & 2.3 & \best{2.8} & \best{1.0} & \best{92.7} & \best{90.0} & \best{96.7} & \best{53.3} & \best{80.0} \\
    \bottomrule[1pt]
  \end{tabular}
\end{table}

\paragraph{Real-world robotic manipulation.}
\label{sec:real}
\textbf{\cref{tab:real}} shows that \ours achieves the best success rate on
all three tasks (\cref{fig:real-robot-main}) with the lowest training compute (2.8\,TFLOPs), inference compute (1.0\,TFLOPs), and latency (92.7\,ms), reaching 90.0\% on Open, 96.7\% on Put, and 53.3\% on
Stack, with an 80.0\% average that exceeds $\pi_{0.5}$ (70.0\%, 5.0\,TFLOPs, 120.4\,ms) by
10.0 percentage points.
Inspecting failure modes across baselines clarifies this gain.
First, action-only VLAs ($\pi_{0.5}$ at 70.0\% and GR00T-N1.6 at 53.3\%, 245.5\,ms) suffer from open-loop drift in multi-stage and contact-rich execution despite robot pretraining: lacking explicit supervision on how actions alter the scene, slight handle-grasp slips in \emph{Open} or millimeter-level misalignments when placing the second and third blocks in \emph{Stack} go uncorrected, toppling the tower or leaving the drawer ajar.
Second, among WAMs, Fast-WAM (50.0\%, 12.0\,TFLOPs train, 143.6\,ms) expends capacity reconstructing full future video frames and overfits to static tabletop appearance and lighting noise under limited real-world demonstrations, causing imprecise grasps, while LaWAM (68.9\%, 198.8\,ms) predicts complete future latents dominated by static scene content and reaches only 36.7\% on \emph{Stack} ($-16.6$ points behind \ours).
By supervising only the residual, \ours filters out static background variations and forces the policy to anticipate fine-grained gripper-object displacements, enabling accurate handle engagement, stable three-block stacking, and reliable drawer closure without external robot pretraining.

\subsection{Residual prediction and ablation analysis}
\label{sec:worldpred}
\label{sec:ablation}

\begin{figure}[!htb]
  \centering
  \includegraphics[width=0.6\linewidth]{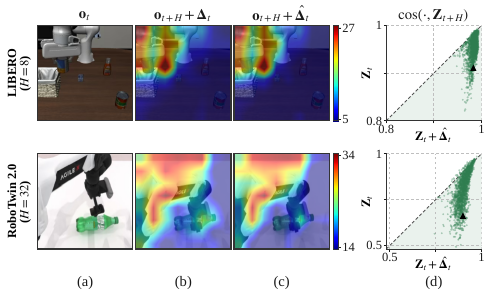}
  \caption{Residual prediction quality on LIBERO (top) and RoboTwin~2.0 (bottom): (a) the current frame $\mathbf o_t$; (b, c) saliency maps of the per-token norm of the true residual $\Res_t$ and the predicted residual $\hat\Res_t$ overlaid on $\mathbf o_{t+H}$ (``$+$''); and (d) cosine similarity to the true future tokens $\Lat_{t+H}$ of the prediction $\Lat_t+\hat\Res_t$ ($x$-axis) versus the current tokens $\Lat_t$ ($y$-axis). The black triangle marks the example window shown in (a)--(c).}
  \label{fig:worldpred}
\end{figure}

\paragraph{Residual prediction quality.}
\textbf{\cref{fig:worldpred}} compares the predicted residual $\hat\Res_t$ with the true residual $\Res_t=\Lat_{t+H}-\Lat_t$ in the DINOv3 token space on LIBERO and RoboTwin~2.0.
In \textbf{\cref{fig:worldpred}(a)--(c)}, $\hat\Res_t$ concentrates on the same patches as $\Res_t$ around the moving arm and manipulated object at steps $t$ and $t+H$, leaving the static background inactive.
Comparing the cosine similarity to $\Lat_{t+H}$ of $\Lat_t+\hat\Res_t$ ($x$-axis) vs.\ $\Lat_t$ ($y$-axis) in \textbf{\cref{fig:worldpred}(d)} shows two findings.
\textit{First}, \textit{$\Lat_t$ and $\Lat_{t+H}$ already share high cosine similarity} (0.960 on LIBERO and 0.823 on RoboTwin~2.0), confirming that scenes are dominated by unchanged background content.
\textit{Second}, \textit{nearly all points lie below the diagonal} (95\% on LIBERO and 91\% on RoboTwin~2.0), as $\Lat_t+\hat\Res_t$ aligns more closely with $\Lat_{t+H}$ (0.982 and 0.899) than $\Lat_t$ alone.

\paragraph{Ablations of the residual branch.}
We ablate our residual branch on RoboTwin~2.0 across three variants:
(1)~\emph{w/o residual branch}, training only the action branch as a pure VLA;
(2)~\emph{future target}, predicting the complete future DINOv3 tokens $\Lat_{t+H}$; and
(3)~\emph{VLM ViT residual}, predicting the feature residual in the VLM's vision tower.

\begin{figure}[!htb]
  \centering
  \includegraphics[width=0.35\linewidth]{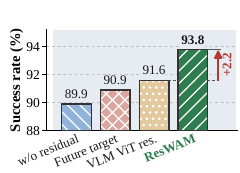}
  \caption{\footnotesize Ablations of residual branch on RoboTwin~2.0 (average success rate, \%).}
  \label{fig:ablation}
\end{figure}

\cref{fig:ablation} reports the average success rates across all 50 tasks, yielding three findings.
\textit{First}, \textit{joint residual prediction is critical for policy performance.} Removing the residual branch (\emph{w/o residual branch}) drops the average success rate from 93.8\% to 89.9\% ($-3.9$ points), confirming the benefit of anticipating scene evolution alongside actions.
\textit{Second}, \textit{learning the residual is substantially more effective than learning the future.} Predicting $\Lat_{t+H}$ (\emph{future target}) reaches 90.9\%, trailing \ours (93.8\%) by 2.9 points and validating \cref{sec:rank,sec:vis}: $\Res_t$ avoids spending capacity on static backgrounds and focuses supervision on moving arms and objects.
\textit{Third}, \textit{frozen DINOv3 tokens provide a superior target space for residual learning.} Predicting the residual in the VLM ViT space (\emph{VLM ViT residual}) reaches 91.6\%, whereas frozen DINOv3 tokens in \ours add $+2.2$ points (93.8\%) due to their finer spatial geometry.

\begin{figure}[!htb]
  \centering
  \includegraphics[width=0.32\linewidth]{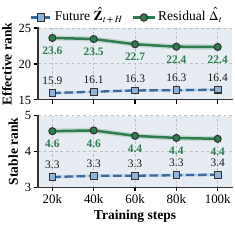}
  \caption{\footnotesize Effective rank (top) and stable rank (bottom) of predicted tokens $\hat\Res_t$ and $\hat\Lat_{t+H}$ across training steps on RoboTwin~2.0.}
  \label{fig:pred-rank-steps}
\end{figure}

\paragraph{Rank of predicted features across training steps.}
To verify whether the spectral advantage of the ground-truth residual (\cref{sec:rank}) transfers to learned predictions, \textbf{\cref{fig:pred-rank-steps}} tracks the effective and stable ranks of the predicted residual $\hat\Res_t$ (\ours) versus the predicted future $\hat\Lat_{t+H}$ (\emph{future target}) from 20k to 100k training steps on RoboTwin~2.0. Throughout training, $\hat\Res_t$ maintains a substantially higher effective rank (22.4--23.6 vs.\ 15.9--16.4) and stable rank (4.4--4.6 vs.\ 3.3--3.4) than $\hat\Lat_{t+H}$, matching the ground-truth residual spectrum and providing richer visual change cues to the action branch via joint attention.

\section{Conclusion}
\label{sec:conclusion}

In this paper, we addressed both policy performance and computational efficiency in world-action models. To improve policy performance, we proposed \emph{visual residual learning} in the space of a frozen DINOv3 encoder. Compared with complete future features, learning visual residuals achieves a higher feature rank across diverse directions of scene change and concentrates spatially on the moving robot arm and manipulated objects rather than static backgrounds, providing a richer and more focused supervision signal. To improve computational efficiency, we designed \ours, a residual world-action model with 2.3B parameters in total, including a lightweight 0.5B dual-branch DiT that jointly denoises actions and frozen DINOv3 visual residuals via conditional flow matching without relying on video generation models or separate world models. Extensive experiments on LIBERO-Plus, RoboTwin~2.0, and RoboDojo show that without robot pretraining, \ours achieves the highest average success rates (81.1\%, 93.8\%, and 9.40\%) alongside the lowest training compute (2.7--4.0\,TFLOPs), inference compute (1.0--1.4\,TFLOPs), and inference latencies (92.2, 99.4, and 99.7\,ms) among all compared VLAs and WAMs. Real-robot evaluations further confirm this advantage: \ours leads every compared method on Open, Put, and Stack at 1.0\,TFLOPs and 92.7\,ms latency, with an 80.0\% average success rate that exceeds $\pi_{0.5}$ by 10.0 percentage points.

\section*{Acknowledgment}
This work was supported by the Cisco Research Award. Chongyu Fan and Sijia Liu were additionally supported in part by the National Science Foundation (NSF) under the CISE Core Program Award IIS-2504263 and the CAREER Award IIS-2338068. We would also like to thank Gengyu Zhang and Yan Yan for helpful feedback on the real-robot implementation.

\bibliography{main}
\bibliographystyle{iclr2025_conference}


\clearpage
\newpage

\clearpage
\appendix
\section*{\Large{Appendix}}
\crefalias{section}{appendix}
\crefalias{subsection}{appendix}
\crefalias{subsubsection}{appendix}
\setcounter{table}{0}
\setcounter{figure}{0}
\renewcommand{\thetable}{A\arabic{table}}
\renewcommand{\theHtable}{A\arabic{table}}
\renewcommand{\thefigure}{A\arabic{figure}}
\renewcommand{\theHfigure}{A\arabic{figure}}

\begingroup
\setlength{\parskip}{3.5pt}
\makeatletter
\renewcommand\section{\@startsection{section}{1}{\z@}{-1.0ex \@plus -0.2ex \@minus -.1ex}{0.4ex \@plus .1ex}{\large\bfseries}}
\renewcommand\paragraph{\@startsection{paragraph}{4}{\z@}{0.3ex \@plus 0.1ex \@minus .1ex}{-0.8em}{\normalsize\bfseries}}
\makeatother

\section{Rank of the Residual and of the Future}
\label{app:rank}

For each window $(t, t+H)$ on LIBERO ($H=8$) and RoboTwin~2.0 ($H=32$), both ranks of \textbf{\cref{eq:ranks}} are computed on the token-centered matrix $\mathbf J\mathbf X$, where $\mathbf X\in\{\Lat_{t+H},\Res_t\}$ denotes either the future tokens or the residual. Centering prevents a component shared by all tokens from counting as a direction; with $N=64$ layer-normalized DINOv3 ViT-B/16~\cite{simeoni2025dinov3} tokens of dimension $D=768$, both ranks are at most $63$.

\section{Projecting the Residual onto the Image}
\label{app:vis}

To visualize the residual in image space (\cref{fig:vis}), we extract layer-normalized patch tokens on the $16\times16$ grid from the frozen DINOv3 ViT-B/16~\cite{simeoni2025dinov3} encoder and compute the per-patch residual $\res_{t,j}=\mathbf z_{t+H,j}-\mathbf z_{t,j}$.
For the \emph{saliency map}, we compute the $\ell_2$ norm $\|\res_{t,j}\|_2$ at each patch, bilinearly upsample it to the image resolution, and blend it with the future observation $\mathbf o_{t+H}$ using the jet colormap.
For the \emph{PCA-RGB map}, we project centered residual tokens onto the top three principal eigenvectors of the residual covariance, normalize each component to $[0,1]$ as an RGB channel, and upsample to the image resolution by nearest-neighbor interpolation.

\section{Training and Implementation Details}
\label{app:training}

\paragraph{Optimization.}
Across benchmarks, we train \ours using AdamW with $(\beta_1,\beta_2)=(0.9,0.95)$, weight decay $10^{-8}$, and BF16 mixed precision with fully sharded data parallelism (FSDP). The learning rate warms up linearly over 1{,}500 steps and then follows a cosine decay schedule from $10^{-4}$ to $10^{-5}$ for the action and residual expert and from $10^{-5}$ to $10^{-6}$ for the VLM backbone. All simulation benchmarks use a global batch size of 256.

\paragraph{LIBERO-Plus setup.}
On LIBERO~\cite{liu2023libero}, we train a single multi-task model on all four suites for 100k steps and evaluate it zero-shot across all 10{,}030 perturbed episodes of LIBERO-Plus~\cite{fei2025libero} with horizon $H=8$ and loss weight $\lambda_w=0.1$. The action is a 7-D end-effector delta with a gripper command, normalized to $[-1,1]$ using the per-dimension 1st and 99th percentiles. At test time, the policy predicts 8 actions and executes the full chunk before replanning.

\paragraph{RoboTwin 2.0 setup.}
On RoboTwin~2.0~\cite{chen2025robotwin}, we train a single multi-task model on the combined clean and domain-randomized demonstrations released with Fast-WAM~\cite{yuan2026fast} with horizon $H=32$ and loss weight $\lambda_w=0.1$. The action is the 14-D absolute joint position normalized by its mean and standard deviation, with binary grippers. At test time, the policy predicts 32 actions and executes 16 steps before replanning.

\paragraph{RoboDojo setup.}
On RoboDojo~\cite{chen2026robodojo}, we train a single multi-task model on the official 35-task simulation training set with horizon $H=32$ and loss weight $\lambda_w=0.1$. The action is the 14-D absolute joint position normalized by its mean and standard deviation. At test time, the policy predicts 32 actions and executes 10 steps before replanning across all 42 tasks (54 evaluation configurations, 2{,}100 episodes).

\paragraph{Real-world setup.}
For the real-world experiments, following the DROID~\cite{khazatsky2024droid} hardware setup, we use a Franka Research 3 (FR3) arm equipped with a Franka Hand on three tabletop manipulation tasks: \emph{Open} (\textit{``Open the top drawer, put the red eraser inside, and close the drawer.''}), \emph{Put} (\textit{``Put the cucumber on the plate.''}), and \emph{Stack} (\textit{``Stack the three blocks into a tower.''}). We train a single multi-task model for 30k steps with a global batch size of 32, horizon $H=32$, and loss weight $\lambda_w=0.1$, using the 8-D end-effector state (3-D position, 4-D quaternion, and 1-D gripper width).

\section{Computational Efficiency}
\label{app:latency}

\paragraph{Training and inference FLOPs protocol.}
The per-sample training (\textit{Train}) and per-chunk inference (\textit{Infer.}) TFLOPs in \cref{tab:libero-plus,tab:robotwin,tab:robodojo,tab:real} are counted directly on the official operator graphs at batch size 1 in BF16 under the camera views, state dimensions, and action horizons of each benchmark. \textit{Infer.} counts all active inference modules per predicted action chunk over 10 denoising steps. \textit{Train} counts algorithmic single-sample compute, comprising one forward pass (including any online frozen encoders or video VAEs) and one backward pass without activation checkpointing.

\paragraph{Inference latency protocol.}
All latencies in \cref{tab:libero-plus,tab:robotwin,tab:robodojo,tab:real} were measured by us
on one NVIDIA RTX A6000. Each model runs its
official inference code in its default precision with batch size 1 and 10
denoising steps under the camera views, state dimensions, and action horizons of each benchmark. The timed
region starts from preprocessed inputs on the GPU and ends with the predicted
actions on the GPU, so CPU-side image resizing, tokenization, and
normalization are excluded. Each model is warmed up for 20 iterations and
then timed over 1{,}000 iterations (100--200 iterations for multi-second video WAMs), measuring wall-clock time with a device
synchronization (\texttt{torch.cuda.synchronize} or JAX
\texttt{block\_until\_ready}) before and after each call. The warmup
excludes any compilation time. \cref{fig:latency-std} reports the per-chunk inference latency ($\mathrm{Mean}\pm\mathrm{Std}$, in ms) across all four benchmarks, where \ours achieves the lowest mean latency (92.2--99.7\,ms).

\begin{figure}[!htb]
  \centering
  \includegraphics[width=\textwidth]{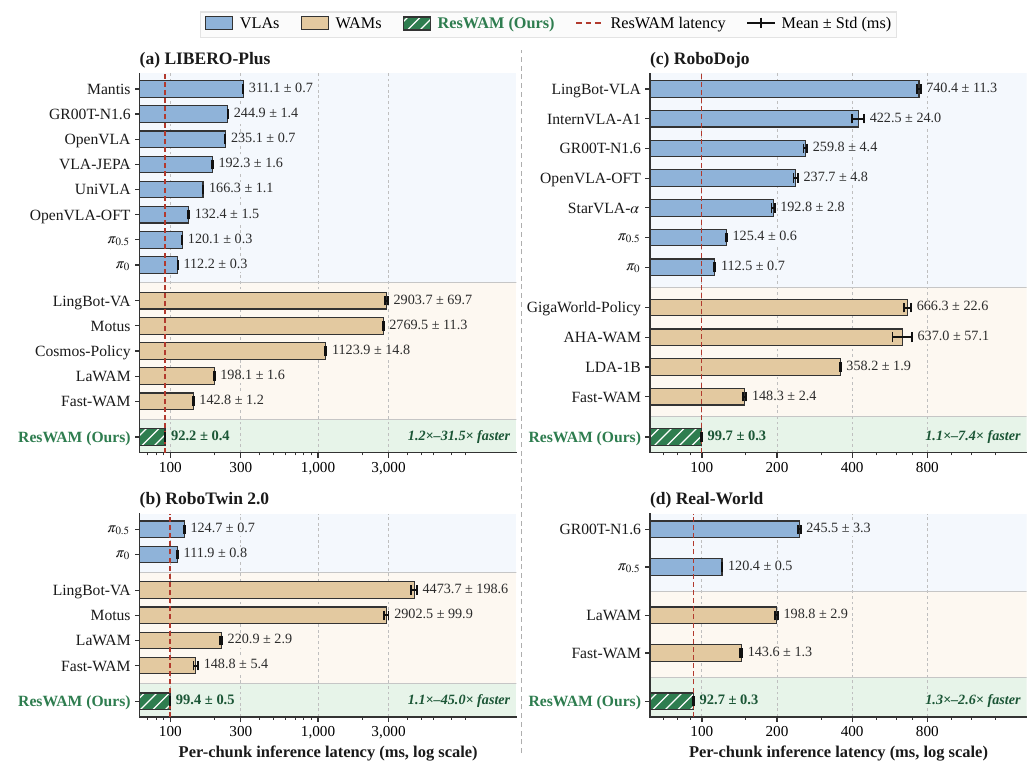}
  \caption{\footnotesize Per-chunk inference latency ($\mathrm{Mean} \pm \mathrm{Std}$, ms) across VLAs, WAMs, and \ours on one NVIDIA RTX~A6000 over (a)~LIBERO-Plus, (b)~RoboTwin~2.0, (c)~RoboDojo, and (d)~Real-World. Horizontal bars and whiskers show sample means and standard deviations on a logarithmic axis, and vertical dashed lines mark the latency of \ours.}
  \label{fig:latency-std}
\end{figure}

\section{Parameter Counts}
\label{app:params}

The parameter counts reported in the Size columns of \cref{tab:libero-plus,tab:robotwin,tab:robodojo,tab:real} are read directly from the tensor headers of the released weights loaded by each method's official inference code, including modules from external releases such as text encoders, VAEs, and frozen vision encoders, with tied or duplicated storages counted once.
\endgroup


\end{document}